\documentclass[letterpaper, 10 pt, conference]{ieeeconf}  

\IEEEoverridecommandlockouts                              
\usepackage{ifpdf}
\usepackage{cite}
\usepackage[pdftex]{graphicx}
\usepackage{amsmath}
\usepackage{amssymb}
\usepackage{algorithm}
\usepackage[noend]{algpseudocode}
\usepackage{array}
\usepackage{subcaption}
\usepackage{url}
\usepackage{multicol}
\usepackage{booktabs}
\usepackage{siunitx}
\usepackage{xcolor}

\title{\LARGE \bf Coordinated Coverage and Fault Tolerance using Fixed-wing Unmanned Aerial Vehicles}

\author{{Sachin~Shriwastav$^{1}$ and Zhuoyuan~Song$^{2}$}%
\thanks{*This work was supported by the College of Engineering of the University of Hawai`i at M\={a}noa.}
\thanks{$^{1}$S.~Shriwastav and $^{2}$Z.~Song are with the Department of Mechanical Engineering, University of Hawai`i at M\={a}noa, Honolulu, HI, 96822 USA}%
\thanks{$^{1}$S.~Shriwastav is with the East-West Center Student Affiliate program.}
\thanks{Emails: \{{\tt\small $^{1}$sachins,$^{2}$zsong}\}  {\tt\small @hawaii.edu}}%
}

\begin{document}

\maketitle
\thispagestyle{empty}
\pagestyle{empty}

\begin{abstract}
This paper presents an approach for deploying and maintaining a fleet of homogeneous fixed-wing  unmanned aerial vehicles (UAVs) for all-time coverage of an area. Two approaches for loiter circle packing have been presented: square and hexagon packing, and the benefits of hexagon packing for minimizing the number of deployed UAVs have been shown. Based on the number of UAVs available and the desired loitering altitude, the proposed algorithm solves an optimization problem to calculate the centres of the loitering circles and the loitering radius for that altitude. The algorithm also incorporates fault recovery capacity in case of simultaneous multiple UAV failures. These failures could form clusters of survivor (active) UAVs over the area with no overall survivor information. The algorithm deploys a super-agent with a larger communication capacity at a higher altitude to recover from the failure. The super-agent collects the information of survivors, and updates the homogeneous radius and the locations of the loitering circles at the same altitude to restore the full coverage. The individual survivor UAVs are then informed and transit to the new loitering circles using Dubin's paths. The relationship with the extent of recoverable loss fractions of the deployed UAVs have been analysed for varying the initial loiter radii. Simulation results have been presented to demonstrate the applicability of the approach and compare the two presented packing approaches.

\end{abstract}

\section{INTRODUCTION}

The utilization of unmanned aerial vehicles (UAVs) for coverage and sensing applications is on the rise with their evolution in terms of speed, endurance, ease of control, and autonomous fleet operation capabilities. UAVs can cover large grounds in short time and provide remote access to information from inaccessible and hazardous areas and environments. Coverage using fixed-wing UAV traditionally means flight cycles over an area through pre-specified flight paths to collect and relay the information to the base for processing. Even though rotor-type UAVs can provide persistent coverage by hovering over an area, they are constrained by their endurance. Fixed-wing UAVs consume significantly less energy to remain airborne for longer duration compared to their rotor-type counterparts~\cite{oettershagen2016long, guo2011development, oettershagen2016perpetual}. Their oftentimes large wing surface areas allow installation of solar panels that may further extend their endurance. Despite being suitable candidates for coverage and sensing applications, fixed-wing UAVs' mobility is typically limited by their minimum cruise speeds and loitering radii for persistent coverage, making the coordination of large fixed-wing UAV fleets challenging~\cite{Song:17a, Song:18d}. This motivates the possibility of the use of fixed-wing UAV fleets for a long term and large scale coverage sensing applications. 

Coverage and sensing applications using UAVs (both fixed-wing and rotor-type) have been a popular research topic with been various theoretical and experimental results. Mozaffari et al.~\cite{ mozaffari2016efficient } uses an efficient deployment of multiple UAVs acting as wireless base stations that provide coverage was analyzed by the ground users. Following this, the downlink coverage probability for UAVs was derived as a function of the altitude and the antenna gain. Next, using circle packing theory, the 3-D locations of the UAVs was determined in a way that the total coverage area was maximized while maximizing the coverage lifetime of the UAVs. In \cite{ varga2015distributed }, teams of fixed-wing micro-aerial vehicles (MAVs) could provide a wide area coverage and relay data in the wireless ad-hoc networks. The authors proposed a distributed control strategy that is based on the attraction and repulsion between MAVs and relies only on local information. Xu et al.~\cite{ xu2011optimal } presented an adaptation of an optimal terrain coverage algorithm for an aerial application. The general strategy involves computing a trajectory through a known environment with obstacles and ensures complete coverage of the terrain while minimizing path repetition. The paper introduced a system that applies and extends this generic algorithm to achieve automated terrain coverage using an aerial vehicle. Danoy et al.~\cite{danoy2015connectivity} presented an online and distributed approach for bi-level flying ad-hoc networks, in which the higher-level fixed-wing fleet serves mainly as a communication bridge for the lower-level fleets that conduct precise information sensing. Chen et al.~\cite{chen2018self} presented a self-organized, distributed and autonomous approach for sensing coverage for multiple UAVs with an approach that takes into account the reciprocity between neighboring UAVs to reduce the oscillation of their trajectories. Nedjati et al.~\cite{ nedjati2016complete } presented a post-earthquake response system for rapid damage assessment. In this system, multiple UAVs are deployed to collect images from an earthquake site and create a response map for extracting useful information. Avellar et al.~\cite{avellar2015multi} presented an algorithm for minimum-time coverage of ground areas using a group of UAVs equipped with image sensors by modeling the area as a graph and solving a mixed integer linear programming problem. Coombes et al.~\cite{coombes2017boustrophedon} addressed the need for an enhanced understanding of the wind effects on fixed-wing aerial surveying, and used Boustrophedon paths based on sweep angle relative to the wind that minimises the flight time. In \cite{coombes2018fixed}, the algorithm took into account environmental factors and aircraft dynamics. By decomposing the complex survey regions into many smaller arrangements, Boustrophedon paths can be used to cover them. Ahmadzadeh et al.~\cite{ahmadzadeh2006multi} presented an algorithm for time-critical cooperative surveillance with a set of unmanned aerial platforms using Integer Programming (IP)-based strategy for feasible trajectories while incorporating the complexity and coupling of the camera fields of view and flight paths. Darbari et al.~\cite{darbari2017dynamic} presented a dynamic path planning algorithm for a UAV surveying a cluttered urban landscape. Voronoi Tessellation of the search space and identification of key waypoints in the form of milestones lead to an efficient mapping of the region to be surveyed. The changes in the environment were handled effectively by the decision process in the form of local or global planner. The application of 3D Dubin's curve lead to smooth and dynamically feasible trajectories at runtime. In \cite{ahmadzadeh2006cooperative}, the authors addressed the generation of team flight plans and controllers that enable a heterogeneous team of UxVs (x: A-Aerial, G-Ground) to maximize spatio-temporal coverage while satisfying hard constraints such as collision avoidance and positional accuracy. Paull et. al~\cite{ paull2013sensor } presented an algorithm where area coverage with an on-board sensor was an important task for a UAV while maintaining an in-situ coverage map based on its actual pose trajectory and making control decisions based on that map. Savla et. al~\cite{savla2007coverage} studied a facility location problem for groups of Dubin’s vehicles,  non-holonomic vehicles constrained to move along planar paths of bounded curvature, without reversing direction. Given a compact region and a group of Dubin's vehicles, the coverage problem is to minimize the worst-case traveling distance.
 
\begin{figure*}
	\centering
	\begin{subfigure}[b]{0.4\textwidth}
	    \includegraphics[width=1\linewidth]{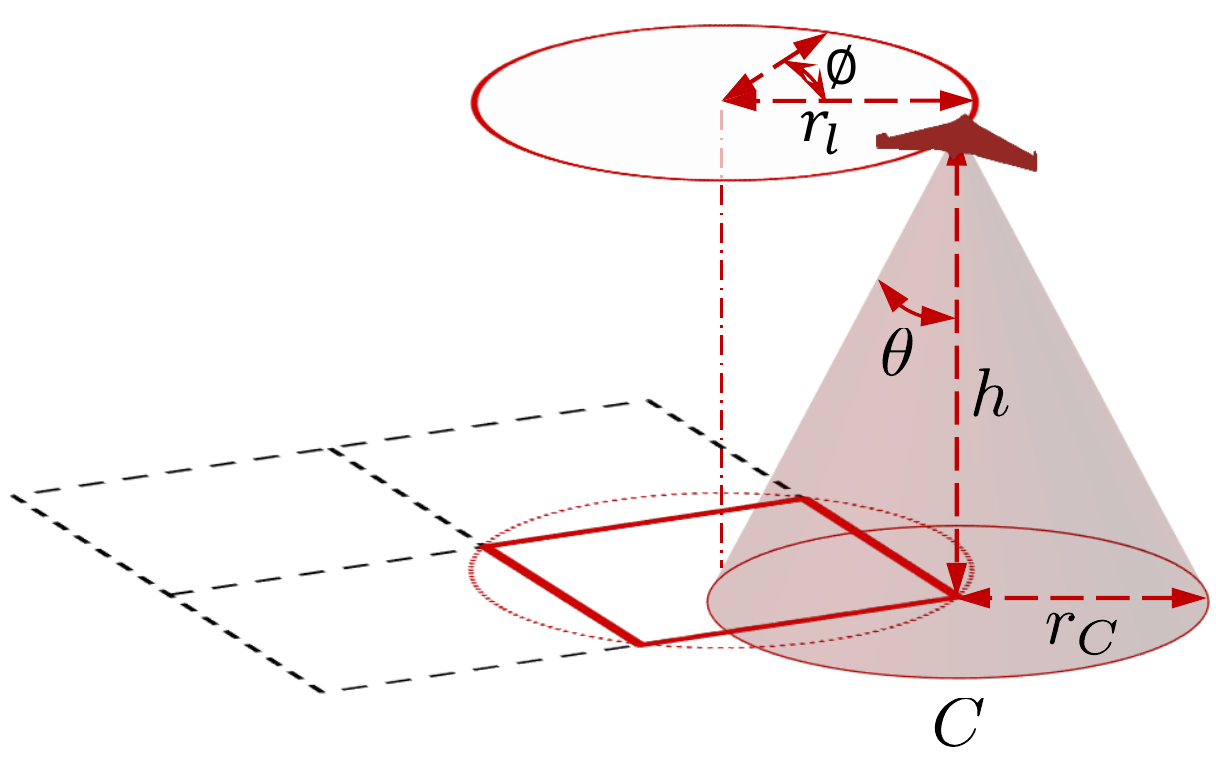}
	    \caption*{(a)}
	\end{subfigure}
	\hspace{8mm}
	\begin{subfigure}[b]{.42\textwidth}
        \includegraphics[width=1\linewidth]{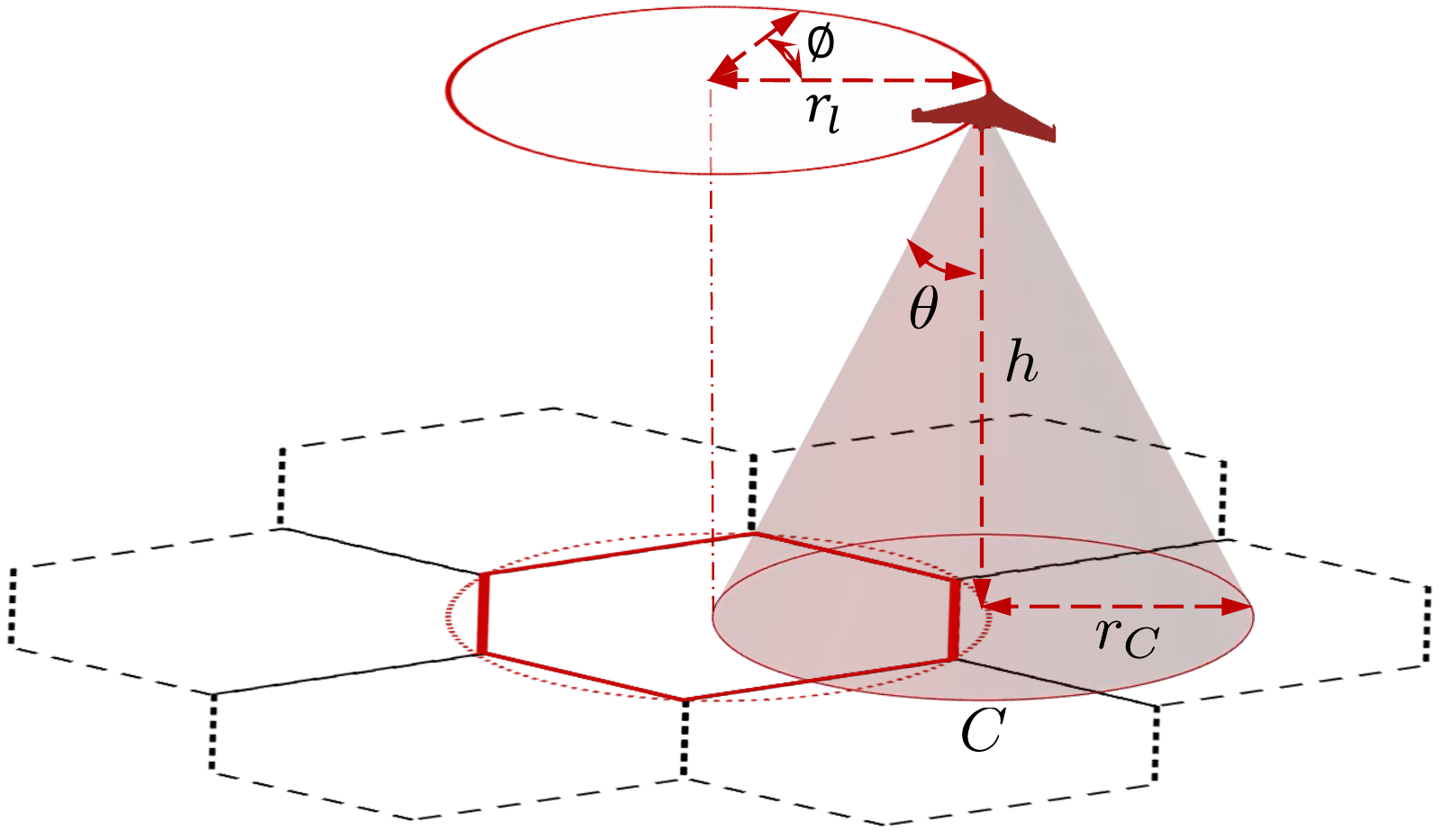}	\caption*{(b)}
	\end{subfigure}
	
    \caption{Parameters for the proposed approach: Efficient packing for full coverage of the area by deployment of synchronised homogeneous UAVs. The loiter circles and their instantaneous phase and coverage footprint is shown for the two presented cases: (a) Square packing; (b) Hexagon packing.}
	\label{fig:illustration}
\end{figure*}

This work presents an algorithm for persistent coverage of an area, by patrolling with a fleet of loitering fixed-wing UAVs at a pre-specified altitude over the area. The loitering circles are packed by inscribing over the packed squares or hexagons as shown in Fig.~\ref{fig:illustration}. In this paper, the circles are packed for both the cases over a rectangular area only. The user is supposed to provide with the number of available UAVs and the desired loitering altitude. It is desired to have enough number of UAVs to be able to start at minimum loitering circle, and have persistent coverage of the area at all times. Based on the available UAV count, the algorithm formulates and solves an optimization problem to compute the centre location of the uniform loitering circles for the given altitude, by maximizing the loitering radius. Following that, Dubin's path algorithm~\cite{savla2007coverage, mclain2014implementing, lugo2014dubins} is used to calculate the deployment paths for the UAVs, from the base to the respective loiter circles over the area to be covered. The paths also take the synchronization of UAVs into account, so that they loiter in the same phase on their respective circles to maximize the effective coverage. The algorithm also incorporates resilience and can handle a failure scenario of simultaneous loss of multiple UAVs. As this type of event could result in clusters of survivors unaware of each other's existence, the need for a global planner arises. On detection of failure, the base deploys a ``super-agent" that has a large communication range and flies at a higher altitude to be able to communicate with all the survivors. The super-agent counts the number of survivors, collects the overall area information, and runs the location optimization algorithm again, to compute the new loiter locations and the updated (larger) loiter radius for the survivors to resume full coverage. It then computes the Dubin's path for the UAVs to transit to their new loitering circles in a synchronized fashion. The UAVs are then informed of their new locations, radius and transition path before they travel to restore full coverage. Although loitering at a larger radius might take away the persistence of coverage, it is ensured that the area is still fully covered in the loitering cycle.

The major contributions of this work are as follows:
\begin{enumerate}
    \item Given a sufficient number of fixed-wing UAVs for an area, the proposed algorithm ensures persistent coverage during the loitering cycles;
    \item The algorithm defines a simple optimization problem for deployment of UAVs over a rectangular area;
    \item The presented algorithm provides coverage resilience by addressing coverage recovery problem in case of simultaneous multiple UAV failures;
    \item With insufficient numbers of UAVs after the failures, the algorithm ensures full coverage of the area using available UAVs;
    \item This work studies and compares the efficiency of hexagon packing over square packing for a given deployment scenario.
\end{enumerate}

The remainder of the paper is organized as follows: Section~\ref{section:prelim} presents the base concept and definitions, Section~\ref{section:approach} discusses the details of the proposed approach, Section~\ref{section:sims} presents the simulation results and discussions. Finally, Section~\ref{section:last} lists some of the possible future work in this domain and concludes the paper.

\section{PRELIMINARIES} 
\label{section:prelim}
The area to be covered by a UAV fleet can be represented as a graph with the vertices representing the locations of the agents (longitude and latitude). The deployed UAVs would represent a different set of nodes $\mathbb{V}'$ as a sub-graph $\mathbb{G}' = (\mathbb{V}',\mathbb{E}')$, and the virtual edges ($\mathbb{E}'$) between the neighboring UAVs represent the active communication link.

\subsection{Definitions}
We define the following quantities to facilitate the presentation of the proposed method. Fig.~\ref{fig:illustration} summarizes these quantities graphically.


\textbf{Field of view (FOV):} The FOV is a physical property of the sensor being used by the UAV and defines the coverage footprint based on the altitude of the platform. In Fig.~\ref{fig:illustration}, The FOV has been marked by $\theta$. Based on the sensor used, the sensing quality ($q$) can be defined as $q \propto \ {1}/{h}$, where $h$ is the loitering altitude of the UAV. This means that the coverage quality decreases linearly as the altitude increases, and vice-versa.

\textbf{Coverage radius ($r_\text{c}$):} The coverage radius is the radius of the coverage footprint of the on-board sensor, given the sensor FOV and the instantaneous height ($h$) of the vehicle.  Coverage radius is directly proportional to the loitering altitude and inversely proportional to the coverage quality, for a given FOV. As seen from Fig.~\ref{fig:illustration}, it is defined as $$r_\text{c} \ = \ h \cdot \tan \theta.$$

\textbf{Loiter radius ($r_\text{l}$):} Any form of a fixed-wing vehicle has constraints on maneuverability and it cannot stay stationary while it is airborne. Instead, it can fly in a circle over the region of interest, called the \textit{loiter circle}. The radius of that circle at a given instant is called the loiter radius. In Fig.~\ref{fig:illustration}, the loiter radius has been shown by $r_{\text{l}}$.

The physical properties and cruising velocity of a fixed-wing UAV system define the lowest value of the loiter radius, called the minimum turning radius~\cite{mclain2014implementing}, given by, $$r_\text{min-turn} \ = \ \frac{v^2}{g} \psi_\text{max},$$ where ${v}\in \mathbb{R}^2$ is the horizontal vehicle velocity, $\psi_\text{max}$ denotes the maximum bank angle and $g$ denotes the gravitational acceleration. It is desired to have $r_\text{l} > r_\text{min-turn}$ to be able to provide coverage while causing less physical strain on the UAV. 

\textbf{Maximum loiter ($r_\text{l-max}$):} This is the maximum loiter radius at which the UAVs can fly, while maintaining full coverage of the area. For hexagon packing, it is given by $r_\text{l-max} \leq r_\text{c}/(\sqrt{3}-1)$. The UAVs can still loiter at circles with radius larger than $r_\text{l-max}$ if necessary at the cost of losing full coverage of the desired area.

\textbf{Communication radius ($r_\text{com}$):} Based on the on-board hardware, a UAV can connect to every other UAV within a certain distance, called the \textit{communication range}. Assuming an isotropic antenna for uniform range, the radius of the coverage is called the communication radius. Its value cannot be less than $\sqrt{2} r_\text{l-max}$ for square packing as the inter-center distance between the loitering UAVs is  $\sqrt{2} r_\text{l}$. The minimum value for hexagon packing is $\sqrt{3} r_\text{l-max}$. It is a different entity from the loitering radius ($r_\text{l}$) and the sensor coverage radius ($r_\text{c}$). For any UAV $k$ at position $x_k$, its neighborhood is defined as, $$\mathcal{N}_k \ \overset{\Delta}{=} \ \{x_i \in \mathbb{V}' \ | \ \text{dist}(x_k, x_i) \ \leq \ r_\text{com} \}.$$

\begin{figure}
    \centering
    \includegraphics[width=0.65\linewidth]{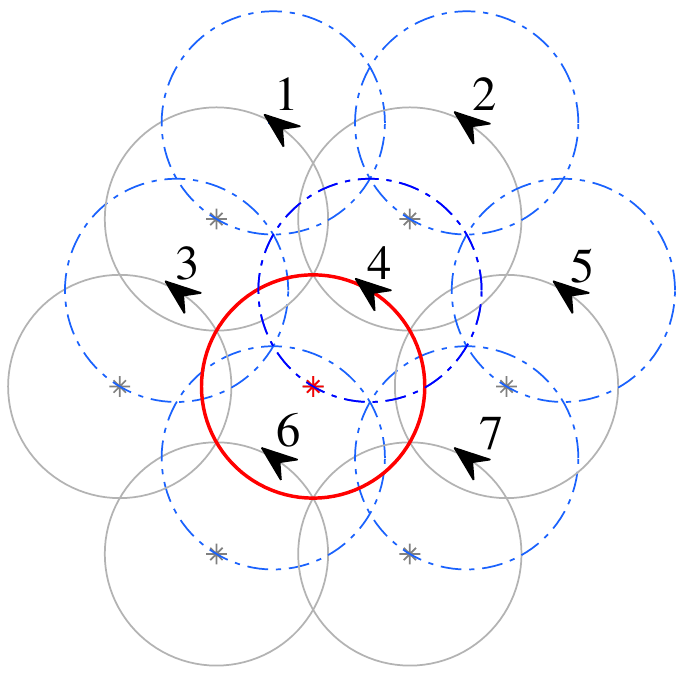}
    \caption{Illustration of persistent coverage in the case of hexagon packing, for $r_\text{l} = r_\text{c}$. Here, the circles in red and grey are the loitering circles (radius $r_\text{l}$), and the blue dashed circles are the instantaneous sensor coverage for each of the loitering circles. The instantaneous position ($\phi = \pi/3$) and heading of the UAVs are shown by arrows. It can be seen that the area under red circle is fully covered at the given instant by UAVs 3, 4, 6 and 7, where UAV 4 is the one loitering over the red circle. This applies to all other circles, as they are persistently covered by their own and neighbor UAVs.}
    \label{fig:persistent}
\end{figure}

\textbf{Persistent coverage:} It is defined as the state when each point in the area is guaranteed to be covered by at least one of the loitering UAVs at every instance of time, throughout the operation period. For hexagon packing, persistent coverage can be maintained if $r_\text{l} \leq  r_\text{c}$ (see Fig.~\ref{fig:persistent}), and the side length of the packing square and hexagon for a given loitering altitude are $r_\text{l}/ \sqrt{2}$ and $r_\text{l}$, respectively.

\textbf{Effective coverage ($E$):} It is defined as the total area covered by a loitering circle within the boundaries of the area of interest, minus the overlap with the immediate neighbors. These overlaps are purposefully introduced to allow the algorithm to cover every point (avoid coverage gaps) in the desired coverage area. The effective coverage for UAV k is defined as, $$E_k \ = \ (1-f) \pi r_l^2 \ - \ \sum_{i \in \mathcal{N}_k} A_{\text{s-}i},$$ where $f$ is the fraction of the circle outside the area of interest and $A_{\text{s-}i}$ (see Fig.~\ref{fig:SectorOverlap}) is its overlap with the neighbor UAV $ i \in \mathcal{N}_{k}$.

\textbf{Full coverage:} It is defined as the state when each point in the area is guaranteed to be covered by at least one of the loitering UAVs at least once every loiter cycle during the operation. For $N$ UAVs deployed in the area $A$, it is achieved when $$A \  \subseteq \ \sum_{i=1}^{N} E_i. $$ This serves as the main objective of the presented work, where we adjust the radius of the loiter circle for the available UAVs to achieve full coverage.

\textbf{Phase synchronization:} For a UAV loitering at an altitude, the phase has been defined in this paper as the angle at which they are. It has been shown in Fig.~\ref{fig:illustration} as $\phi$. We assume that all the loitering UAVs have the same phase at every instant of time for maximum separation, and hence the largest effective coverage.

\textbf{Super-agent:} This is an agent with enhanced communication and computation capability, which is used as a global planner in case of simultaneous multiple node failures.


\subsection{Assumptions}
The following assumptions are made to simplify the analysis:
\begin{enumerate}
\item The UAVs are homogeneous, that is, they have the same size, weight, minimum turn radius, communication and sensing capabilities;
\item The cruising velocity is constant and uniform for all the UAVs;
\item The UAVs always fly at the same altitude specified by the user, even with reduced fleet size;
\item Each UAV knows its location at any point in time;
\item The UAVs are automatically able to communicate with any other UAV within its communication range ($r_\text{com}$).
\end{enumerate}

These assumptions are for ease of analysis and initial verification of the proposed approach. Homogeneous UAVs allow for simpler calculations because of the same dynamics. The algorithm can focus on other application aspects because of this assumption. The algorithm can be adapted for heterogeneous UAVs in future. The altitude is kept constant to keep the coverage quality constant as for instance, the sensing quality of a sensor (for example, camera) is directly proportional to the altitude of the UAV platform. Changing the altitude will give a rise to the need of a new analysis metric as the coverage quality would change. Also, the UAVs are often equipped with efficient inertial measurement unit (IMU) and global positioning system (GPS) sensors for accurate location, altitude and orientation information (accurate to few centimeters). For real life situations, the assumptions like same cruising velocity, always synchronized phase, lag-free communication may pose obstacles like collisions and data package drops. Relaxing these assumptions will serve to make the algorithm more suited to practical applications, versatile, and scalable, which is among the future scope of this work.

\section{PROPOSED APPROACH}
\label{section:approach}
The details of the proposed approach are presented in this section. Fig.~\ref{fig:illustration} shows the basic set up for the approach. In Fig.~\ref{fig:illustration}(a), the UAV is shown loitering at an altitude $h$ over a square packed area, along the loiter circle with radius $r_\text{l}$ with instantaneous coverage footprint marked by $r_\text{c}$. The sensor FOV for the given altitude has been marked by $\theta$. Fig.~\ref{fig:illustration}(b) shows the equivalent setup and parameters  for the hexagonal packing. The proposed algorithm deploys the UAV fleet over the area with either of those packing methods and handles the cases of simultaneous multiple UAV failures, as summarized in Algorithm~\ref{alg1}. The upper bound on the run time of this algorithm is $\mathcal{O}(N)$, for a network of $N$ UAVs. The details of the approach have been discussed below. 

\subsection{Initial Deployment}
This phase of the algorithm deals with the initial deployment of the UAVs in the area, based on the available UAV count and loitering altitude, by using location optimization technique to calculate the loiter radius and the locations. It is preferred to have sufficient number of UAVs to be able to deploy them at the smaller loiter radius ($r_\text{l} \leq r_\text{c}$), to have persistent coverage and some redundancy to recover from node failures.  This phase starts with the user providing the parameters (area information and the UAV count) and terminates when the UAVs are deployed in the area. The two proposed packing methods, and the location optimization are discussed below.

\subsubsection{Square Packing}
In this case, the homogeneous loitering circles of the radius ($r_\text{l}$) calculated based on the available UAV count inscribe the squares with side length $ \sqrt{2} r_\text{l}$ packed in the area. This case is uniform and thus has the same number of squares in all rows. If a full square leaves some area uncovered near the boundary in either of the X- and Y-directions, an additional square is placed which lies fractionally outside the desired area. This fractionally inside square is assigned to one UAV to have full coverage. The packing starts with the center for the first square being placed at $(r_\text{l} \cdot \cos (\pi /4), \ r_\text{l} \cdot \sin (\pi /4))$, and then other centres are placed at distance $\sqrt{2} r_\text{l} $ distance in X-direction. For the remaining rows, the X-coordinates of the first row can be copied while adding $\sqrt{2} r_\text{l} $ to the Y-coordinate in each step in Y-direction till the rectangle boundary is covered. This is visualised in Fig.~\ref{fig:Sim_HexSq_Initial}(a). The values of this overlap and other parameters for the hexagon packing approach are listed in Table~\ref{table:HexVsSq}. Since the inter-center distance is less than hexagon packing for this method, the inter-circle overlaps are larger and hence the resultant minimum effective area is less compared to the hexagon packing approach.


\begin{figure}
    \centering
    \includegraphics[width=0.65\linewidth]{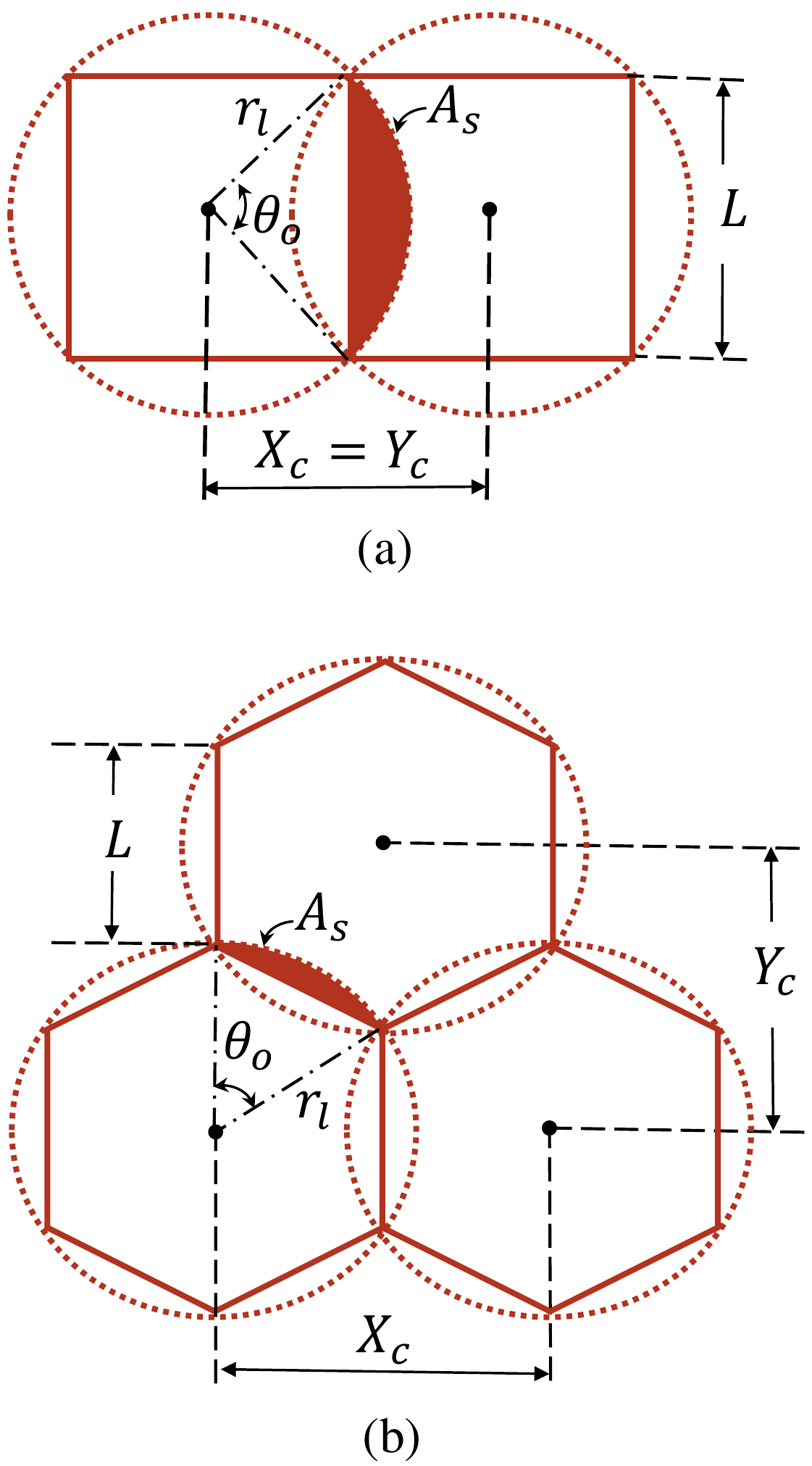}
    \caption{Illustration of the measurement parameters and overlap between two neighbouring loitering circles for (a) Square packing (b) Hexagon packing. Here, $L$ is the length of the side of the inscribed polygon, $X_\text{c}$ and $Y_\text{c}$ are the distances between the adjacent centers in X- and Y- directions respectively, $\theta_\text{o}$ is the sector overlap angle, $A_\text{s}$ (shaded in brown) is the overlap area with a neighbor, and $r_\text{l}$ is the loiter radius.}
    \label{fig:SectorOverlap}
\end{figure}

\subsubsection{Hexagon Packing}
In this case, the area is packed with uniform hexagons with the side length equal to the loitering radius ($r_\text{l}$) calculated by solving the following optimization problem. The loitering circles inscribe these hexagons and have uniform radius. As shown in Fig.~\ref{fig:SectorOverlap}(b) and listed in Table~\ref{table:HexVsSq}, the inter-center distances between two neighboring uniform hexagons are $\sqrt{3} r_\text{l} $ and $3r_\text{l}/2$ in the X- and Y- directions, respectively. This case is not uniform like square packing and the number of UAVs alternates between two values for alternate rows, even for a rectangular area, as seen in Fig.~\ref{fig:Sim_HexSq_Initial}(a). The placement of an additional UAV for a fractionally uncovered area is done here as well, in both directions, as required. The packing starts from one of the vertices of the rectangular area, which is chosen as the origin. The first center is placed at a distance $(r_\text{l} \cdot \cos (\pi /6), \ r_\text{l} \cdot \sin (\pi /6))$ from the origin and then placed along the X-direction at distances $\sqrt{3} r_\text{l} $. For the second row, the center starts at the line $x=0$ for the rectangle boundary at the height $3r_\text{l} /2$ from the first row, and then continued along the X-direction similarly. These two rows are then alternately distance mapped, till the Y-direction boundary is covered. This can be visualised in Fig.~\ref{fig:Sim_HexSq_Initial}(a). The aim of this approach is to achieve full coverage while minimizing the overlap between the loitering circles of neighboring UAVs. The additional circular sector for each hexagon (marked in solid in Fig.\ref{fig:SectorOverlap}(b)) is half of the overlap area with the neighboring circle in that direction. The values of this overlap and other parameters for the hexagon packing approach are listed in Table~\ref{table:HexVsSq}.


\subsubsection{Location Optimization for Hexagon Packing}
Even though it is desirable to start with enough UAVs to deploy with a radius smaller than $r_\text{c}$, this might not always be possible. In addition, as the simultaneous multiple node failure scenario occurs, the remaining UAVs will lose the persistent coverage and loiter at larger radius for full coverage, covering each point in the area at a time interval of $2 \pi r_\text{l} /v$. To achieve this, an optimization problem is formulated and solved for each failure scenario. This optimization is also necessary for the initial deployment if the UAV count is not enough to fly at $r_\text{l} \leq r_\text{c}$. This optimization uses the available number of UAVs ($N$), X-limit of the rectangular area ($X_\text{area}$), Y-limit of the area ($Y_\text{area}$) as inputs and provides the radius value ($r_\text{l}$) and number of UAVs to be deployed in X-direction rows and Y-direction columns, $n_\text{x}$ and $n_\text{y}$ respectively. An initial guess is to be provided for the desired output parameters, which will be refined over iterations. The optimization terminates if the optimal value of $r_\text{l}$ is obtained for the given set of input parameters. If $X = [r_\text{l} \ n_\text{x} \ n_\text{y}]^T$ is the desired output vector and $\sigma = [\sigma_1 \ \sigma_2 \ \sigma_3]^T$ is the tuning parameter, the optimization problem can be formulated as follows:

\begin{equation*}
\begin{aligned}
\underset{r_\text{l}}{\text{minimize}} \qquad & \frac{\sigma_1}{r_\text{l}^2}, \\
\text{subject to} \qquad & \sqrt{3} (n_\text{x} - 1) \ \leq \ X_\text{area} \leq \ \sqrt{3} n_\text{x}, \\
& \frac{3r}{2} (n_\text{y} - 1) \ \leq \ Y_\text{area} \leq \ \frac{3r}{2} n_\text{y}, \\
& n_\text{x} + n_\text{y} + \lfloor \frac{n_\text{y}}{2} \rfloor = N. \\
\end{aligned}
\end{equation*}
The floor operator, $\lfloor * \rfloor $, is used to accommodate the possible difference in number of UAVs in alternate rows. The initial deployment for square packing is a relatively simpler problem as the number of UAVs to be deployed in each row is the same, and the total count is simply $n_\text{x} \cdot n_\text{y}$. Since we focus on the efficiency of the hexagon packing over the square packing, so the details of initial deployment for square packing will not be discussed separately. 


\begin{table}
	\caption{Comparison of parameters between hexagon and square packing; (see Fig.~\ref{fig:SectorOverlap}).}
	\centering
	\begin{tabular}{|c|r|r|}
	    \hline 
	    Parameter & Square packing & Hexagon packing \\
		\hline 
		$L$ & $r_\text{l} \sqrt{2}$ & $r_\text{l}$  \\ 
		\hline 
		$X_\text{c}$   & $\sqrt{2} r_\text{l} $ & $\sqrt{3} r_\text{l} $  \\
		\hline 
	    $Y_\text{c}$  & $\sqrt{2} r_\text{l} $ &  $3r_\text{l}/2$ \\
	    \hline 
		$\theta_\text{o}$  & $\pi/2$ & $\pi/3$ \\ 
		\hline 
		$A_\text{s}$ & $ (\pi-2)r_\text{l}^2/ 4$ & $ (\pi-3)r_\text{l}^2/6$ \\
		\hline 
		$E$  & $ (4-\pi)r_\text{l}^2$ & $ (6-\pi)r_\text{l}^2$ \\
		\hline 
	\end{tabular}
	\label{table:HexVsSq}
\end{table}

\begin{algorithm}[t!]
	\caption{Hexagon packing: Deployment and recovery}

	\begin{algorithmic}[1]
		\Statex \textbf{/* Initial Deployment */}
		\Statex \textbf{Input:} $N, \ X_{\text{area}}$, and $Y_{\text{area}}$
		\Statex \text{Output:} Loiter radius and center coordinates of packing hexagons in the area
		\State $X = [r_\text{l}, \ n_\text{x},\ n_\text{y}]^T \text{, desired output vector}$ \label{rep1}
		\Statex Solve Optimization problem in Section.~\ref{section:approach}-A-3
		\State Get the desired vector values \label{rep2}
		\State Plot a first row center at $(r_\text{l} \cdot \cos (\pi/6), \ r_\text{l} \cdot \sin (\pi/6))$ \label{rep3}
		\State Plot second row center at $x=0$, and $3r_\text{l}/2$ above first row\
		\While {New Center $ < X_{\text{area}}$}
		    \State Plot centers in X-direction for both rows
		\EndWhile
		\While {Rows count $ < Y_{\text{area}} / n_\text{y}$}
		    \State Distance map Row-1 \& 2 alternately in Y-direction
		\EndWhile 
		\State Find Dubin's path to each loiter circle from current location (Base, for initial deployment) 
		\State Deploy fixed-wing UAVs at given altitude $h$ \label{rep4}
		\Statex \textbf{/* Failure Detection */}
		\If {$dist(i,j) \ \leq  \sqrt{3} r_\text{l}$} \label{rep5}
		    \State UAV$_i$ and UAV$_j$ are connected
		    \If{ UAV$_i$ cannot connect to UAV$_j \ \forall \ i \in \mathcal{N}_{j}$}
		    \State UAV$_j$ dropped out
		    \EndIf
		\EndIf
		\State Base receives the failure message 
		\Statex \textbf{/* Failure Handling (Recovery) */}
		\State Base deploys the super-agent
		\State Super-agent compiles $N_\text{new}$ and the location information
		\If {$N_{\text{new}}>0$}
		    \State Repeat lines \ref{rep1} to \ref{rep2} for calculating $r_\text{l-new}$ for $N_\text{new}$
		\EndIf
        \If {$r_{\text{l-new}} \leq r_\text{c}$}
            \State Repeat lines \ref{rep3} to \ref{rep4} for calculating new centers
		\Else
	          \State Recovery not possible
		\EndIf
        \If{New centers calculated}
		    \State Find Dubin's path to each new loiter circle 
		    \State Move UAVs  and full coverage restored
		\Else
		    \State Recovery failed
		\EndIf \label{rep6}
		\State Repeat lines \ref{rep5} to \ref{rep6} for every failure instance
	\end{algorithmic}
	\label{alg1}
\end{algorithm}

\begin{figure*}
	\centering
	\begin{subfigure}[b]{0.35\textwidth}
	    \includegraphics[width=1\linewidth]{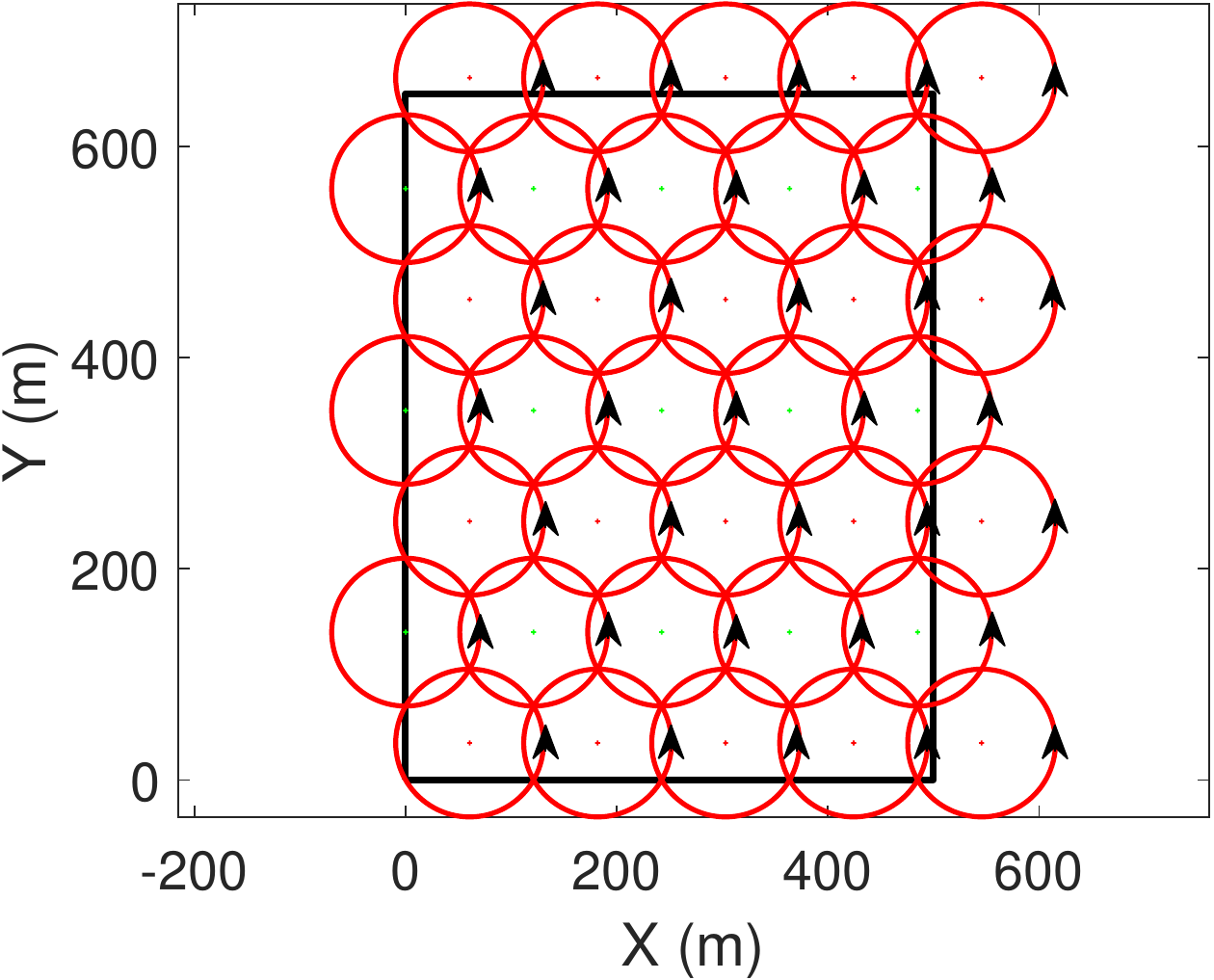}
	    \caption*{(a)}
	\end{subfigure}
	\hspace{8mm}
	\begin{subfigure}[b]{.35\textwidth}
        \includegraphics[width=1\linewidth]{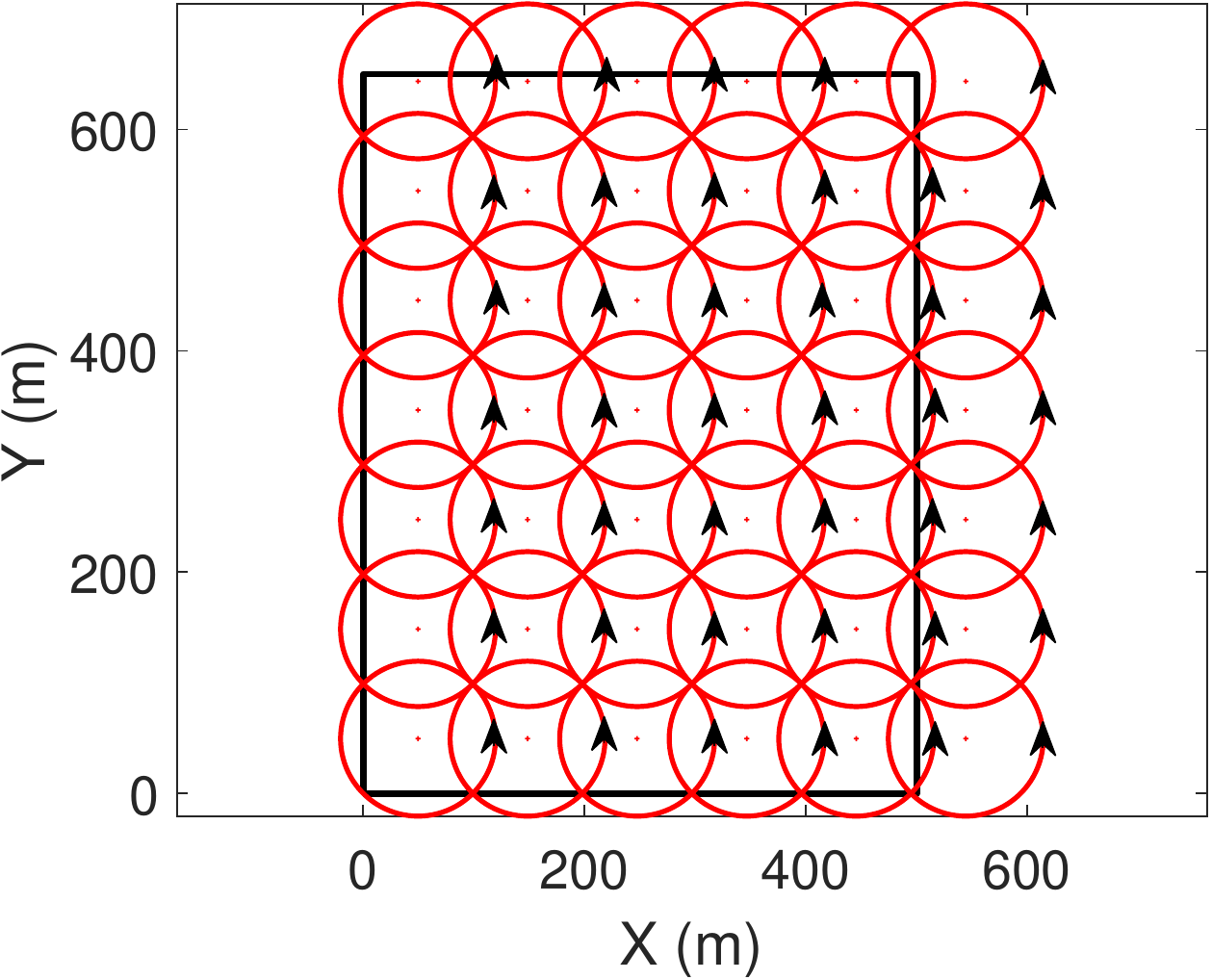}	
        \caption*{(b)}
	\end{subfigure}
	
    \caption{Simulation results to show how the hexagon packing is efficient compared to the square packing, and requires less number of UAVs for the same value of loiter radius over the same area: (a) Hexagonal packing (b) Square packing. The instantaneous phase and direction of the loitering UAVs are shown by black arrowheads in both case.}
	\label{fig:Sim_HexSq_Initial}
\end{figure*}

It can be noted from the above discussion and Table~\ref{table:HexVsSq} on the two presented packing approaches that hexagon packing has less overlap, and hence higher minimum effective area, by a margin of $2 r_\text{l}^2$. This shows that hexagon packing is more effective and requires less number of UAVs to cover the same area for the same loiter radius (see Fig.~\ref{fig:Sim_HexSq_Initial}). Since most of the steps and computations are similar for both the approaches except for the numbers, the rest of the discussion in this paper will be based on hexagon packing approach.

\subsubsection{Deploying the UAVs}
After the centres of the loiter circles for the given number of nodes and the loiter radius for the given altitude are available, Dubin's path algorithm~\cite{savla2007coverage, mclain2014implementing, lugo2014dubins} can be used to calculate the deployment paths for the UAVs from the base to the respective loiter circles over the area. The UAVs then get deployed and loiter over the area, providing full coverage.

\subsection{Failure Detection}
There are many possible reasons for systems like these to fail. Failures could occur due to external impacts (e.g., blast), UAV instrument failure, power source failure, or many other possible reasons. The UAV is considered `unrecoverable' after the failure. Failure detection is a local phenomenon, when one or more agents suddenly drop out of operation. The immediate neighbors detect the absence of their neighbor and pass the failure message to the base. In case no agent directly connected to the base survives, the base detects the failure by itself. In square and hexagon packing, a UAV can be connected to up to four and six other UAVs, respectively, for $ r_\text{l} \leq r_\text{com}$. This number can be larger if the communication range is larger, depending on the application. If a UAV drops out, all other UAVs directly connected to it detect the failure. Based on the active communication link, each UAV maintains a list of the neighbors' state with all `1's. For instance, in hexagon packing, if a UAV drops out, its neighbors change the respective label to '0', indicating its failure. That is, for UAV$_k$ with six neighbors in the hexagon packing, $\mathcal{N}_k^\text{state} = [1, \ 1, \ 1, \ 1, \ 1, \ 1]$, for neighbors $\mathcal{N}_k = [\mathcal{N}_{k1}, \ \mathcal{N}_{k2}, \ \mathcal{N}_{k3}, \ \mathcal{N}_{k4}, \ \mathcal{N}_{k5}, \ \mathcal{N}_{k6}]$ means all-active neighbors and operations. If neighbor $\mathcal{N}_{k3}$ drops out, the list is updated to $\mathcal{N}_k^\text{state} = [1, \ 1, \ 0, \ 1, \ 1, \ 1]$. For a simultaneous multiple node loss scenario that leaves survivor clusters over the area, the UAVs at the border of the cluster who have lost their immediate neighbor detect the mass failure. However, these clusters are unaware of any other survivor clusters over the area, so a locally distributed recovery process is not feasible. This leads to the need for a global planner, which is served by the super-agent. The recovery process after the multiple node failures will be discussed next.

\subsection{Failure Handling (Recovery)}
The most important objective of the proposed algorithm is to provide the full coverage. It is intuitive that the UAVs have to fly on larger loiter circles to restore the coverage, but the trade off is that the system loses the persistent coverage if it cannot deploy UAVs to loiter at $r_\text{l} \leq r_\text{c}$. As the UAVs start loitering at $r_\text{l} > r_\text{c}$, it can guarantee that each underlying point gets covered at least once in a loiter cycle. When the number of available UAVs is not enough for persistent coverage, the algorithm shifts its objective to obtain full coverage using the available UAVs.

As discussed previously, there is a need for a global planner for recovery in this case since there is a possibility with no information on survivors available on a global scale. The base thus deploys a super-agent after the failure detection, which is capable of communicating at a larger range. The super-agent flies into the area at a higher altitude, and receives the information on all possible survivor clusters spread all over the area. Once this phase is over, it is solely responsible for generating the optimal recovery decision, efficient in terms of recovery time and distance travelled. As summarized in Algorithm~\ref{alg1}, it first counts the survivor UAVs and then compiles the information. If no UAV has survived in the area, the recovery fails and the super-agent returns to the base. Next step is to check if there are enough survivor UAVs to recover the full area, given the constraint on the coverage footprint radius for the loiter altitude. The algorithm does not consider flying at a higher altitude to keep the coverage quality constant. Based on the number of survivors $N_\text{new}$ and available information on area boundary $X_\text{area}$ and $Y_\text{area}$, it then performs the optimization discussed in Section \ref{section:approach}-A-3 to calculate the new loiter radius, $r_\text{l-new}$. If $r_\text{l-new} \leq r_\text{c}$ for the given altitude, persistent coverage can be restored, and if $r_\text{c} < r_\text{l-new} \leq (\sqrt{3}-1)r_\text{c}$, the algorithm is able to restore full coverage. Otherwise, the super-agent notifies the base of the deficit and it is up to the base to resupply UAVs, lose coverage and continue, or terminate the operation. If $r_\text{l-new}$ is in the permissible range, it now computes the centers of the new loitering circles with a larger radius. The super-agent then computes their paths to the new loiter circles using Dubin's path algorithm, with an additional consideration of collision avoidance and phase synchronisation in the new set up. The survivor UAVs are then informed of their new assignment and the super-agent returns to the base with a 'recovery successful' message. The UAVs then break out of their current loiter circles, follow the calculated Dubin's paths, and start loitering at the new assigned locations, to fully restore the coverage. 

If the area is too large for a super-agent to communicate and navigate, and thus poses computational burden on the super-agent, the algorithm can be modified to deploy more than one super-agent with pre-defined area jurisdictions. These super-agents can then collect the information from respective sub-areas and coordinate among themselves to restore the full coverage using the survivor UAVs. This is beyond the scope of the presented work. 

\begin{figure*}
	\centering
	\begin{subfigure}[b]{0.31\textwidth}
	    \includegraphics[width=1\linewidth]{figs/ICUAS_Hex70_Initial.pdf}
	    \caption*{(a)}
	\end{subfigure}
	\hspace{3mm}
	\begin{subfigure}[b]{.31\textwidth}
        \includegraphics[width=1\linewidth]{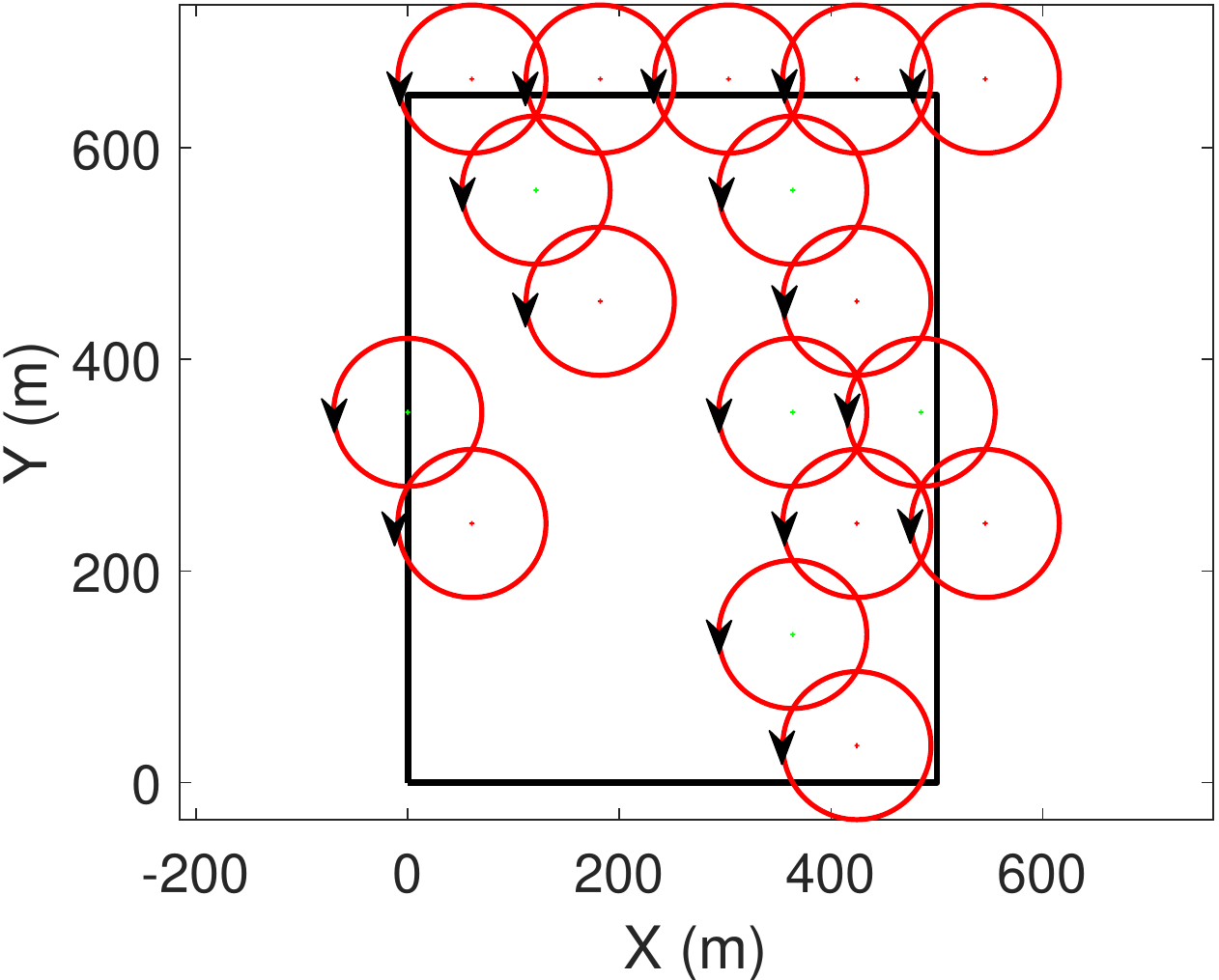}	
        \caption*{(b)}
	\end{subfigure}
	\hspace{3mm}
	\begin{subfigure}[b]{.31\textwidth}
        \includegraphics[width=1\linewidth]{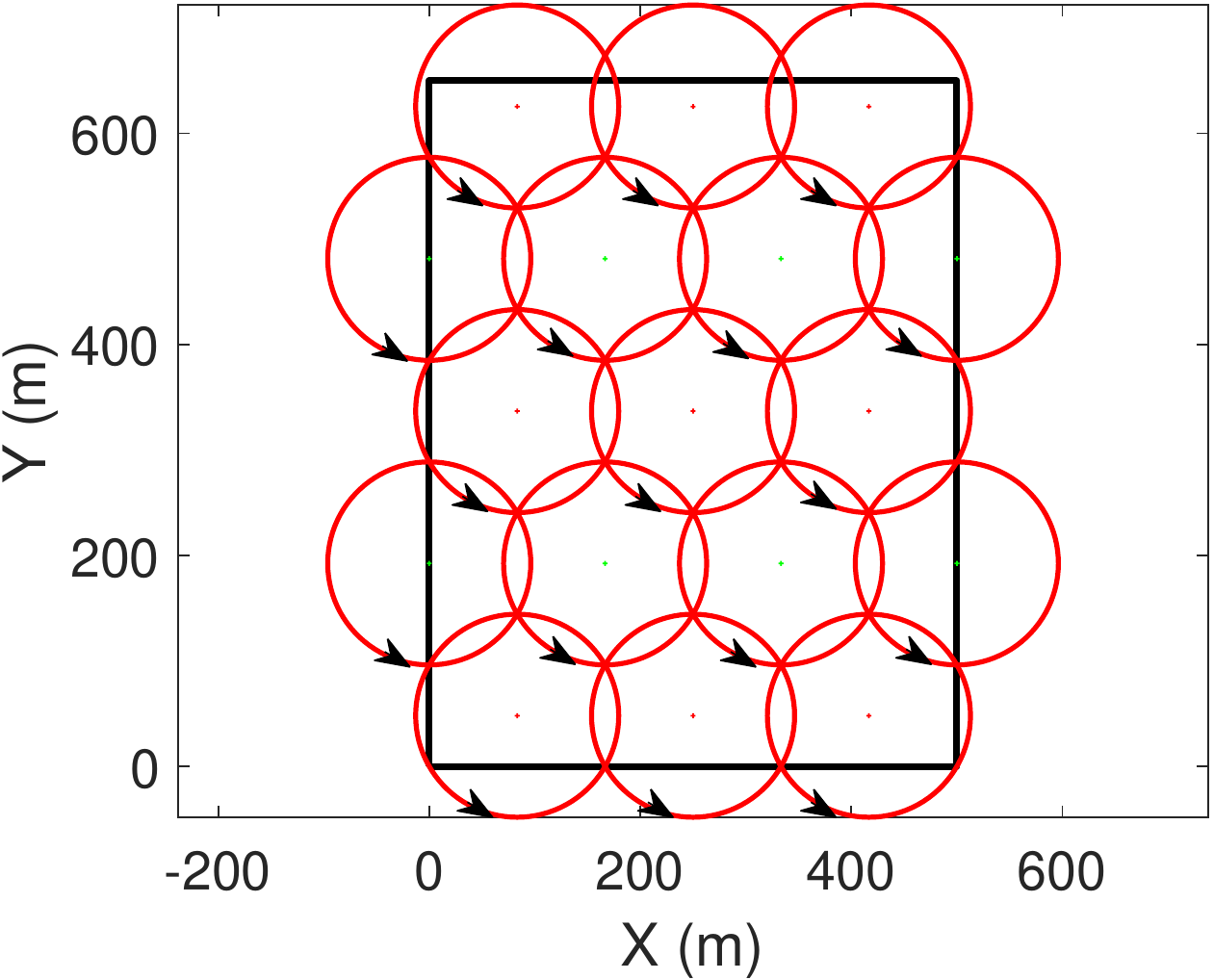}	
        \caption*{(b)}
	\end{subfigure}
    \caption{Simulation of the proposed approach for fault tolerance after simultaneous multiple agent drop out to maintain full coverage: (a) Initial deployed network with full coverage; (b) Simultaneous multiple node loss over the deployment area resulting in clusters; (c) Recovery of full coverage by re-optimizing the loiter circles' location and the loiter radius based on the number of survivors over the network. The instantaneous phase and direction of the loitering UAVs are shown by black arrowheads in each case.}
	\label{fig:Simulation}
\end{figure*}

In this approach, all the survivor UAVs have to break off from their current loiter circles to trace a Dubin's path to the loiter circle of a larger radius at the same altitude, and join in at the prescribed point and phase. This can be achieved by controlling the break-off point, the join-in point, and the headings at both points. Typically, Dubin's paths are created as a combination of circular sections and straight lines, with an aim to minimize the travel time and distance. The motion of the UAV is constrained into six options: straight, left turn, right turn, helix left turn, helix right turn, and no motion. The  equations of motion and the generation of Dubin's paths are well-explored topics of the existing literature~\cite{savla2007coverage, mclain2014implementing, lugo2014dubins}. For multiple UAVs, the most important consideration is not to have more than one UAV at a point during the transit. The paths are thus calculated for individual UAVs, ensuring that they do not collide with any other UAVs.

\section{SIMULATION RESULTS}
\label{section:sims}
To verify the applicability, the proposed algorithm was applied to various area sizes, while controlling the number of UAVs and loiter circle radius. Major simulation results, along with parametric comparisons are discussed below.

\begin{table}
	\caption{Parameters used in the simulation case shown in Figs.~\ref{fig:Simulation} and \ref{fig:SqVsHex_Initial}. Bottom half of the table shows the data for Fig.~\ref{fig:Simulation} for the sample application of the algorithm.}
	\centering
	\begin{tabular}{|l|r|}
		\hline 
		$X_\text{area}$ (m) & 500  \\ 
		\hline 
		$Y_\text{area}$ (m) & 650  \\
		\hline 
	    $r_\text{l-max}$ (m) & 100 \\
		\hline 
		$r_\text{l}$ (m) & [50 60 70 80 90] \\ 
		\hline 
		\hline
		Number of Initial Nodes & 35 ($r_\text{l}$  = 70) \\
		\hline 
		Number of Lost Nodes & 18 ($\approx$50\%) \\
		\hline 
		Number of Survivors & 17  \\
		\hline
		$r_\text{l-new}$ (m) & 96.22 \\
		\hline 
	\end{tabular}
	\label{table:simdata}
\end{table}

\subsection{Application on an area}
The simulation was carried out for various scenarios by changing area dimensions, initial number of UAVs and initial loiter radius ($r_\text{l}$). Table~\ref{table:simdata} lists the simulation parameters for Fig.~\ref{fig:Simulation}, which shows the case for $r_\text{l} = 70$ metres. In these figures, the deployment area has been marked by a black rectangle and each red circle represents the loiter path for a fixed-wing UAV. As shown in Fig.~\ref{fig:Simulation}(a), the area is initially covered by UAVs loitering at $r_\text{l}$ providing full coverage. It is to be noted that the algorithm implies additional UAVs to be deployed, even for a small fraction of the uncovered desired area, to fulfill its primary objective of full coverage. In Fig.~\ref{fig:Simulation}(b), a random simultaneous multiple node loss scenario was applied. This scenario randomly chose and wiped out over half of the UAVs from the area, resulting in two survivor clusters. It can be seen from the figure that the cluster of two UAVs would not have any information about the larger cluster and vice-versa. For lack of global information, neither of them are able to make optimal re-deployment decision. On detection of this failure event, the base deploys a super-agent (not shown in the picture) that flies to the centre of the area, loiters around there and collects the information of both the survivor clusters. Following that, it solves the optimization for ($N_\text{new} = 17, \ X_\text{area}, \ Y_\text{area}$) and calculates $r_\text{l-new}$ to be 96.22 metres. As $r_\text{l-new}$ is still less than the $r_\text{l-max}$ value for the given set up, the super-agent decides that the coverage can be fully recovered. It then calculates the centres for the 17 new loiter circles with this new $r_\text{l-new}$ to fully cover the area, assigns one survivor UAV to each of them, and passes on the decision to the individuals. The super-agent also computes the Dubin's path for each of the survivor UAVs to their new loiter locations, while keeping collision avoidance and phase synchronization in account. The role of the super-agent ends there. The survivor UAVs then follow their respective paths to move to the new locations and restore the coverage. The updated loiter circles with the fully restored coverage are shown in Fig.~\ref{fig:Simulation}(c).

\begin{figure}
    \centering
    \includegraphics[width=0.65\linewidth]{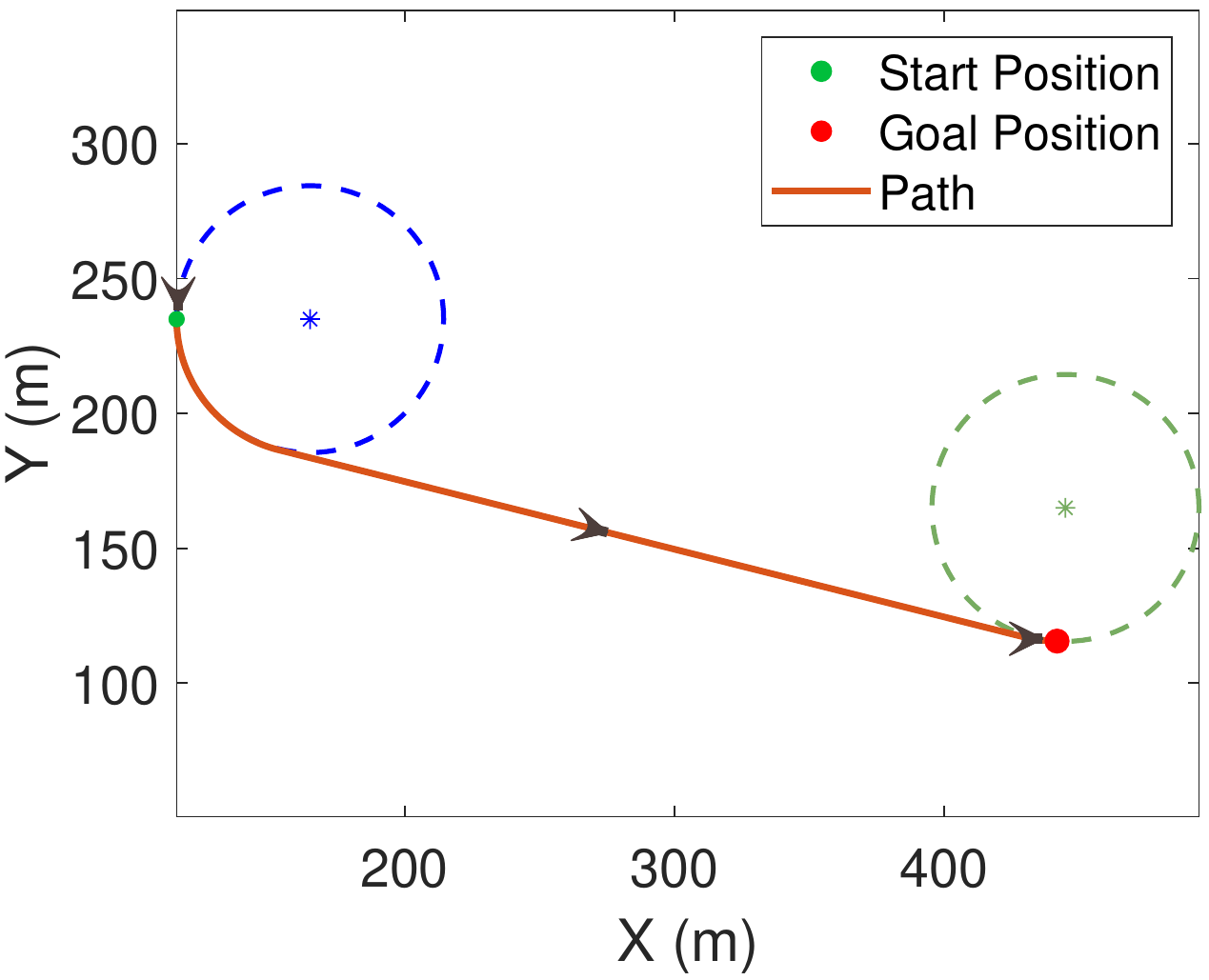}
    \caption{Illustration of the transition path for a UAV while transiting to a new assigned loiter circle (from blue to green), using Dubin's path algorithm. The black arrowheads denote the instantaneous flight direction.}
    \label{fig:dubin}
\end{figure}

\begin{figure}
    \centering
    \includegraphics[width=0.6\linewidth]{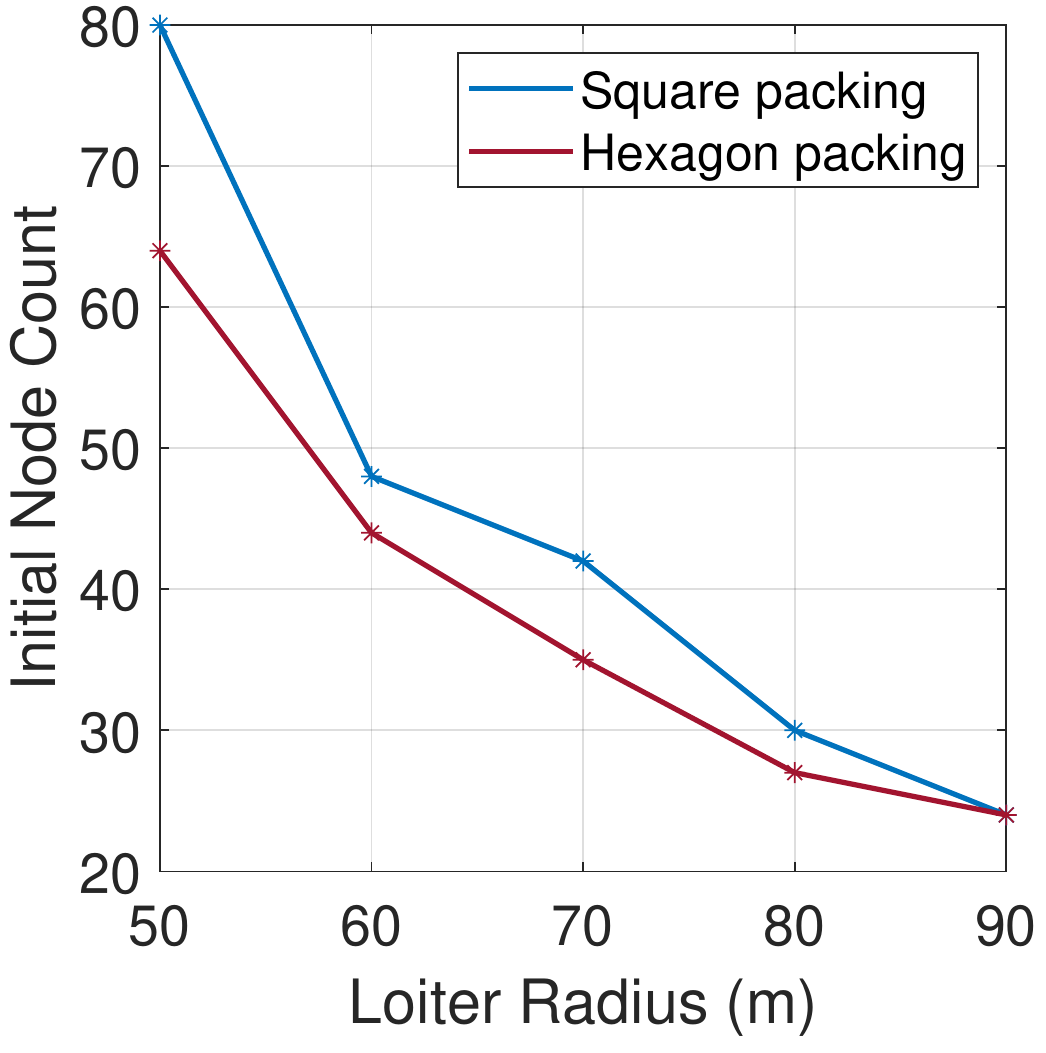}
    \caption{Comparison of number of initial deployed UAVs over the same area for square and hexagon packing for different loiter radius values.}
    \label{fig:SqVsHex_Initial}
\end{figure}

\begin{figure*}
	\centering
	\begin{subfigure}[b]{0.25\textwidth}
	    \includegraphics[width=1\linewidth]{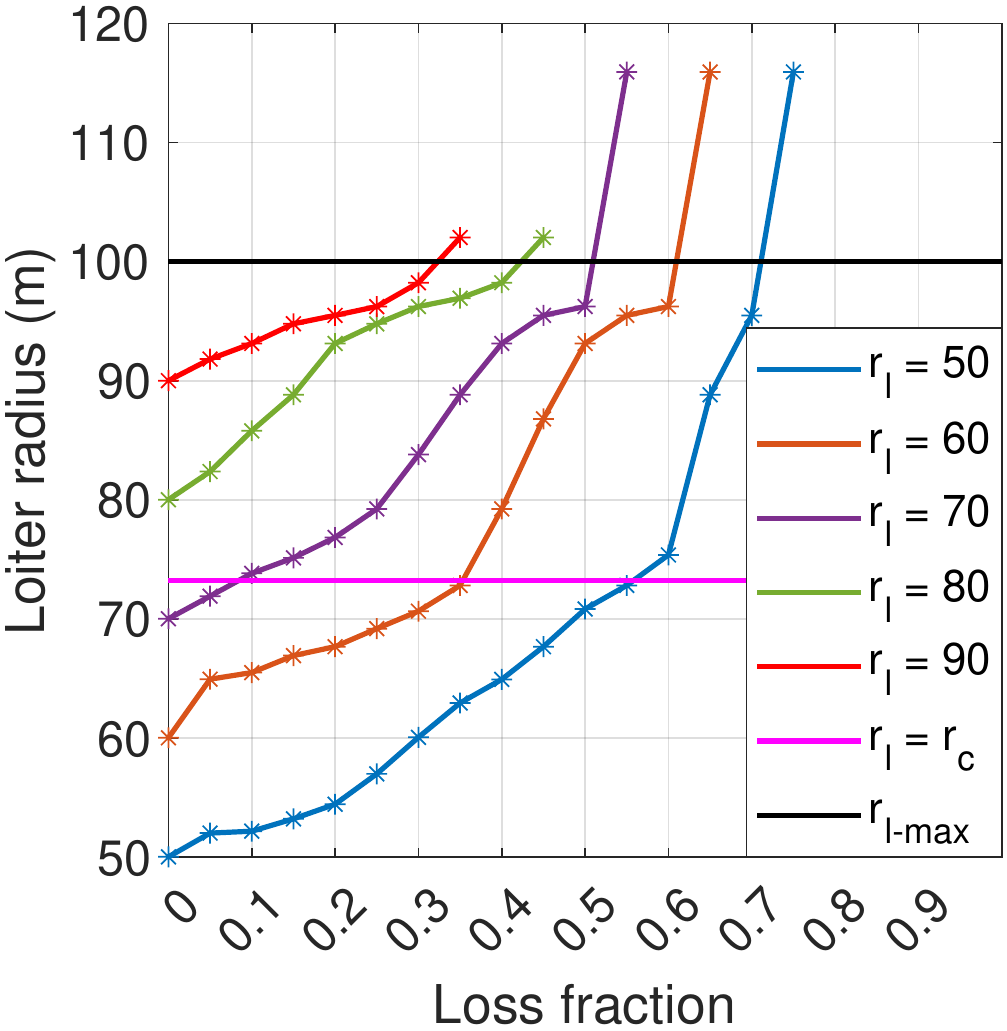}
	    \caption*{(a)}
	\end{subfigure}
	\hspace{8mm}
	\begin{subfigure}[b]{.25\textwidth}
        \includegraphics[width=1\linewidth]{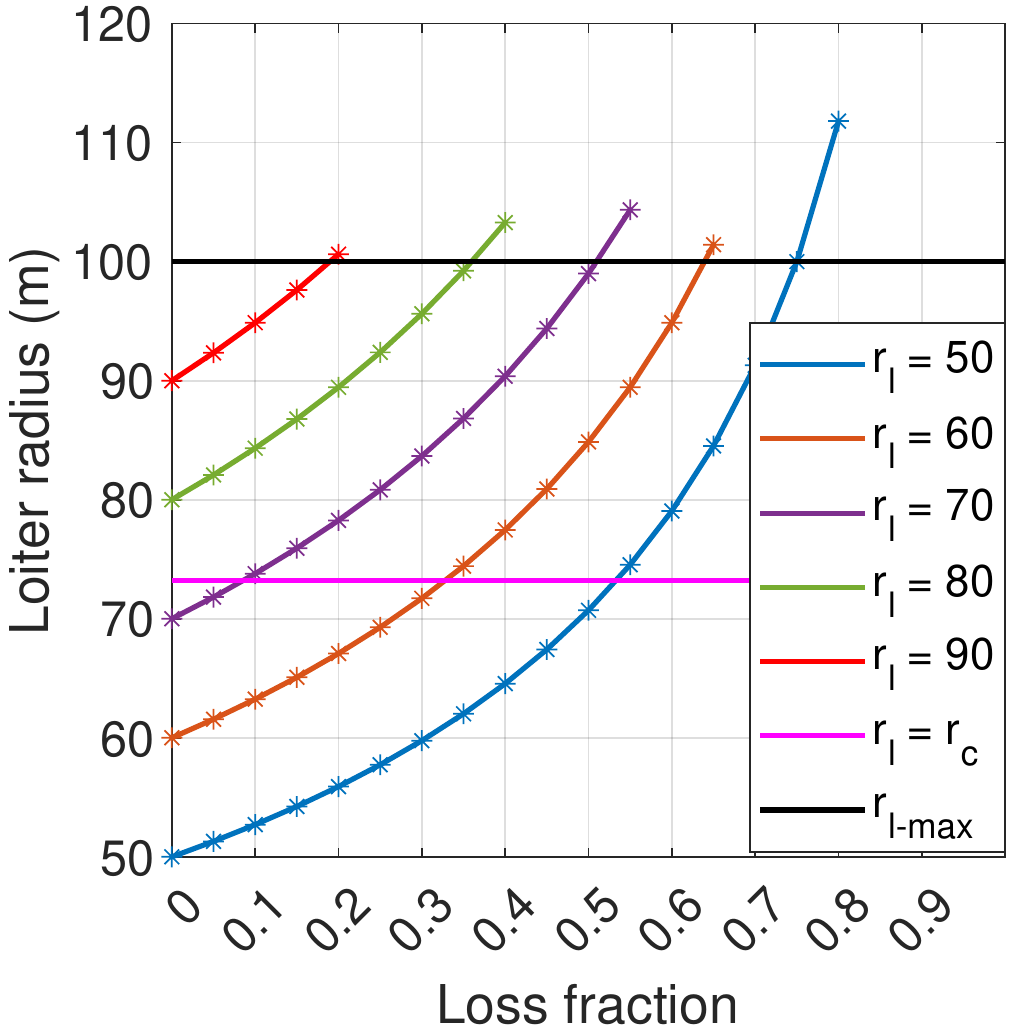}	\caption*{(b)}
	\end{subfigure}
	
    \caption{Comparison of the proposed approach performance with the ideal case (no overlap, boundaries exact multiple of the radius value) in terms of change in loiter radius caused by the fraction of nodes lost, for various values of initial loiter radius: (a) Proposed approach (b) Ideal case. The figures show the $r_\text{l}$ values for which the network will lose its persistent and full coverage abilities.}
	\label{fig:LossCases_SimVsIdeal}
\end{figure*}

Fig.~\ref{fig:dubin} shows a sample Dubin's path for transition of a UAV, to provide an insight of how it is applied. The UAV breaks off from the blue loiter circle at the point marked in green, and traces the path in the solid red curve to reach the green loiter circle, joining in at the point marked in red. The instantaneous headings are shown in the figure. The UAV then uses loiters in the green circle. It is to be noted that the path length for multiple UAVs will be different, to maintain phase synchronisation after moving to the new location.

\subsection{Comparison}
We compare the initial deployment results of the square and the hexagon packing. Figs.~\ref{fig:Sim_HexSq_Initial}(a) and (b) show the initial deployment plot for $r_\text{l} = 70$ metres over the same area, which is marked as black rectangle. Each circle represents the loiter path for a UAV. Both the approaches successfully pack the area, but hexagon packing uses a smaller number of UAVs (35) compared to the square packing (42). This is mostly due to the extent of overlap between the neighboring loiter circles. It is visibly apparent that the inter-circle overlap is higher in the square packing approach, reducing its minimum effective coverage, which causes it to deploy more UAVs. This is in agreement with the theoretical analysis in Section~\ref{section:approach}-A and Table~\ref{table:HexVsSq}. 
Adding a layer to this comparison, Fig.~\ref{fig:SqVsHex_Initial} presents the equivalent result for multiple values of $r_\text{l}$. It can be seen that the square packing deploys a larger number of circles for all values of $r_\text{l}$. However, the difference narrows down for the larger radii while deploying over the same area, which is mostly due to decreasing number of circles and hence overlaps, and also due to nearing the largest coverage radius without losing coverage. It can be concluded that hexagon packing is efficient compared to the square packing.

Fig.~\ref{fig:LossCases_SimVsIdeal} presents the plot of loss fraction against loiter radius, for different values of initial loiter radius, that is, the portion of initially deployed nodes each case can lose and maintain the persistent coverage, or still fully recover, before starting to lose coverage. Fig.~\ref{fig:LossCases_SimVsIdeal}(a) and (b) present the simulation results and ideal case respectively. Unlike the simulation case, the ideal case considers no overlap. In both the plots, curves for each starting loiter radius have been marked in various colors and labeled. The magenta line shows the point where the UAV network loses persistent coverage ability ($r_\text{l}= r_\text{c}$) and the black line shows the maximum allowed loiter radius ($r_\text{l-max}$), for the given altitude. It is basically the cut off point as the UAVs start to lose coverage of the internal area of their loiter circle beyond that value of $r_\text{l-max}$. To start with a certain radius value means that there are enough number of UAVs available to be deployed to fully cover the desired area at that particular value of $r_\text{l}$. As seen in the figure, for example, when the initial deployment starts at $r_\text{l}= 70$ metres (purple line in Fig.~\ref{fig:LossCases_SimVsIdeal}(a)), the UAVs have to start loitering at newly assigned circles with radius 73.81 meters after losing 10\% of the initially deployed UAVs, and they cannot continue persistent coverage. The full coverage can still be restored. The simulation results are still satisfactory compared to the ideal case, as the cut off values for loss fraction to start losing coverage are in the close vicinity of the ideal case values. The inconsistencies in the simulation curves are caused by the fraction loiter circles which appear outside the boundaries, and need to be rounded to next full circle. One interesting fact to note is that the network can fully recover the coverage even after a loss of over 70\% of the initial deployed nodes, for the starting radius $r_\text{l} = 50$ metres at the given altitude.

\section{CONCLUSIONS}
\label{section:last}
Using fixed-wing UAV for sensing and coverage applications is an evolving field, with the complexity and extent of application being on the rise. This paper presents an approach to deploy a fleet of UAVs over an area to achieve full coverage at all times for a long term, while using the minimum number of UAVs. Two approaches for packing the area were discussed and compared: square packing and hexagon packing, and hexagon packing proved to be superior because of less inter-circle overlap within the area between the neighbors. The initial deployment implements the proposed approaches and solves an optimization problem for the optimal loitering radius for a given number of UAVs. The algorithm also incorporates resilience in the UAV network, which can recover from simultaneous node loss to fully recover the coverage. The proposed recovery approach considers scenario of isolated clusters of survivors and utilizes an external super-agent to make the recovery decision, which involves relocating the survivors to new optimized locations at the same altitude, and making them loiter at a larger radius to fully recover the area with the reduced fleet size. Simulation results have been presented to verify the applicability of approach and show its efficacy.

There are a number of future directions for this work, including the experimental verification of the proposed algorithm, considering practical scenarios such as collision avoidance, asynchronized phase, communication lag and more. The constraint on shape of the deployment area can be lifted, which will make it more suited to real-life geographical applications. Another research direction is to make the algorithm adaptive for heterogeneous UAVs, and to introduce weights on the deployment area based on coverage information importance.


\bibliographystyle{IEEEtran}
\bibliography{IEEEabrv,ran}

\end{document}